\documentclass[letterpaper]{article} % DO NOT CHANGE THIS
\usepackage{aaai2026}  % DO NOT CHANGE THIS
\usepackage{times}  % DO NOT CHANGE THIS
\usepackage{helvet}  % DO NOT CHANGE THIS
\usepackage{courier}  % DO NOT CHANGE THIS
\usepackage[hyphens]{url}  % DO NOT CHANGE THIS
\usepackage{graphicx} % DO NOT CHANGE THIS
\usepackage{natbib}  % DO NOT CHANGE THIS AND DO NOT ADD ANY OPTIONS TO IT
\usepackage{caption} % DO NOT CHANGE THIS AND DO NOT ADD ANY OPTIONS TO IT
\usepackage{algorithm}
\usepackage{algorithmic}
\usepackage{amsmath}
\usepackage{booktabs}
\usepackage{multirow}
\usepackage{xcolor,colortbl}
\usepackage{amssymb}
\usepackage{cuted}

\usepackage{newfloat}
\usepackage{listings}

\usepackage{newfloat}
\usepackage{listings}
\DeclareCaptionStyle{ruled}{labelfont=normalfont,labelsep=colon,strut=off} % DO NOT CHANGE THIS
\floatstyle{ruled}
\newfloat{listing}{tb}{lst}{}
\floatname{listing}{Listing}
\title{DiffusionLane: Diffusion Model for Lane Detection}
\author{
    Kunyang Zhou\textsuperscript{\rm 1,\rm 2},Yeqin Shao\textsuperscript{\rm 1}\thanks{Correspond author.}\\
}
\affiliations{
    \textsuperscript{\rm 1}School of ZhangJian, Nantong University \\ 
    \textsuperscript{\rm 2}School of Automation, Southeast University\\
    \{\texttt{kunyangzhou@seu.edu.cn}, \texttt{hnsyk@ntu.edu.cn}\}
}

\usepackage{bibentry}
\begin{document}

\maketitle

\begin{abstract}
In this paper, we present a novel diffusion-based model for lane detection, called DiffusionLane, which treats the lane detection task as a denoising diffusion process in the parameter space of the lane. 
Firstly, we add the Gaussian noise to the parameters (the starting point and the angle) of ground truth lanes to obtain noisy lane anchors, and the model learns to refine the noisy lane anchors in a progressive way to obtain the target lanes. 
Secondly, we propose a hybrid decoding strategy to address the poor feature representation of the encoder, resulting from the noisy lane anchors. Specifically, we design a hybrid diffusion decoder to combine global-level and local-level decoders for high-quality lane anchors. 
Then, to improve the feature representation of the encoder, we employ an auxiliary head in the training stage to adopt the learnable lane anchors for enriching the supervision on the encoder.
% Experimental results show that DiffusionLane possesses a strong generalization ability and favorable performance
Experimental results on four benchmarks, Carlane, Tusimple, CULane, and LLAMAS, show that DiffusionLane possesses a strong generalization ability and promising detection performance
compared to the previous state-of-the-art methods. For example, DiffusionLane with ResNet18 surpasses the existing methods by at least 1\% accuracy on the domain adaptation dataset Carlane. 
Besides, DiffusionLane with MobileNetV4 gets 81.32\% F1 score on CULane, 96.89\% accuracy on Tusimple with ResNet34, and 97.59\% F1 score on LLAMAS with ResNet101. Code will be available at \url{https://github.com/zkyntu/UnLanedet}.
\end{abstract}

% Uncomment the following to link to your code, datasets, an extended version or similar.
%
% \begin{links}
%     \link{Code}{https://aaai.org/example/code}
%     \link{Datasets}{https://aaai.org/example/datasets}
%     \link{Extended version}{https://aaai.org/example/extended-version}
% \end{links}

\section{Introduction}
Lane detection is a fundamental task in computer vision and autonomous driving, playing a crucial role in adaptive cruise control and lane keeping. It aims to predict the location of lanes in the given image.
Existing lane detection methods can be divided into three categories: segmentation-based~\cite{pan2018spatial,zheng2021resa}, anchor-based~\cite{zheng2022clrnet,honda2024clrernet,xiao2023adnet}, and parameter-based methods~\cite{tabelini2021polylanenet,feng2022rethinking}.
Among the existing methods, anchor-based methods achieve excellent performance via predefining the high-quality lane anchors, attracting wide attention. Most anchor-based approaches adopt the learnable anchors to fit the distribution of the dataset. Although this way brings a high performance on the specific dataset,
it suffers from poor generalization ability in the distribution-shift scenarios and requires re-training in these scenarios, damaging the convenience of the model.

To address this issue, we reformulate the anchor-based methods as a denosing diffusion process, i.e., directly predicting lanes from a set of random lane anchors. Starting from the random lane anchors without any learnable parameters,
we expect to gradually refine the random lane anchors to target lanes. This \textit{noise-to-lane} approach does not require the learnable lane anchors, avoiding overfitting on the training dataset.

Diffusion model~\cite{song2020denoising,blattmann2023stable}, a prominent class of generative models for image synthesis, generates images through an iterative denoising process. This process is viewed as \textit{noise-to-image} process.
Since the diffusion model displays a strong generalization ability, subsequent researches explore the diffusion model in perception tasks such as object detection~\cite{chen2023diffusiondet}, object tracking~\cite{xie2024diffusiontrack}, and image segmentation~\cite{ji2023ddp}. However,
to the best of our knowledge, there is no prior work that successfully explores the diffusion model in lane detection.

\begin{figure}[t!]
    \centering
    \includegraphics[width=0.45\textwidth]{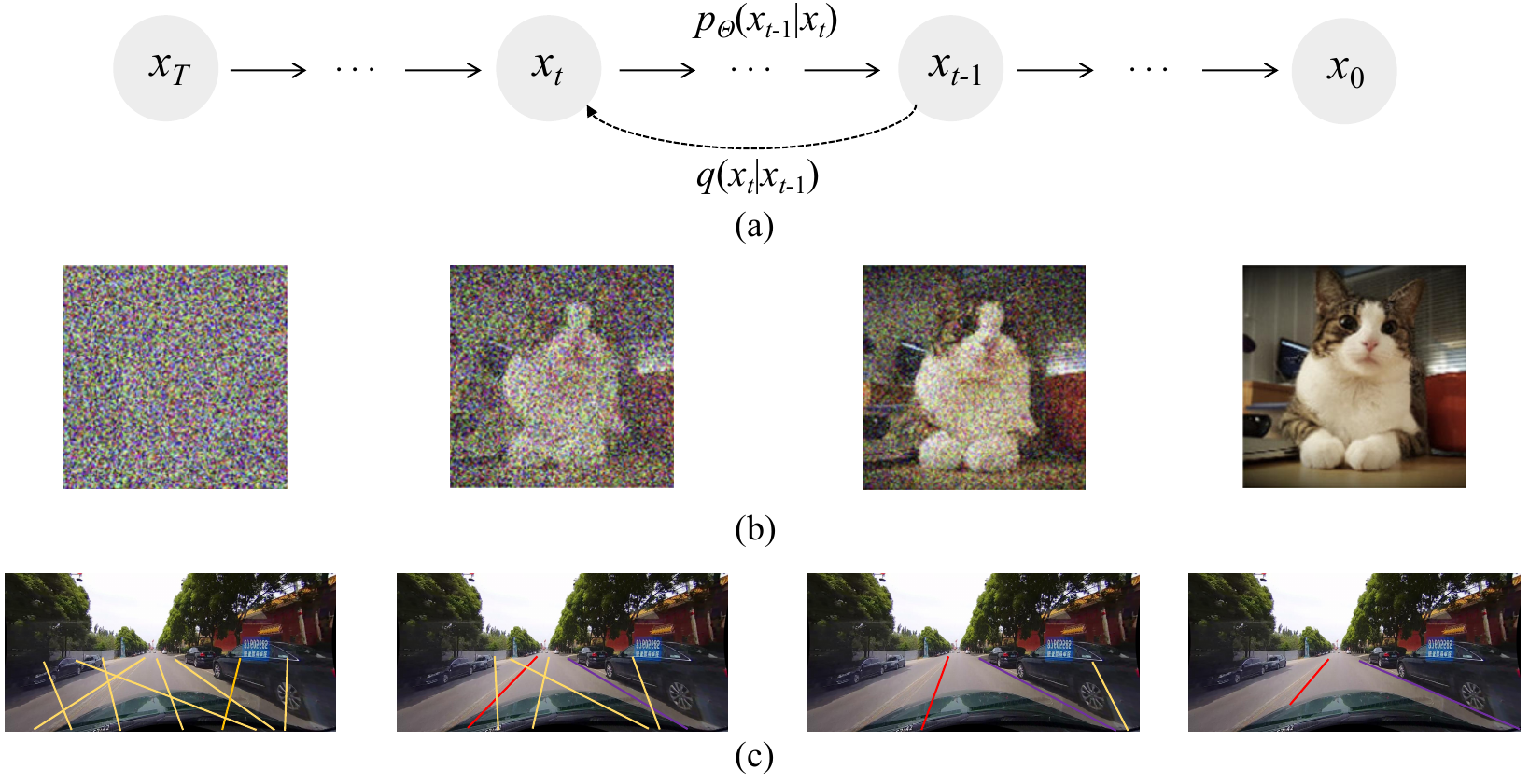}
        \caption{Diffusion model for lane detection. (a) The diffusion model where $q$ denotes the diffusion process and $p_{\theta}$ represents the reverse process. (b) \textit{Noise-to-image} process from noisy pixels to target image. (c) Our \textit{noise-to-lane} paradigm from noisy lane anchors to target lanes.}
    \label{fig:1}
    % \vspace{-0.5cm}
\end{figure}

In this paper, we propose DiffusionLane, casting the lane detection task as a denosing diffusion paradigm over the space of the starting point and the angle of the lane. As shown in Figure~\ref{fig:1}, our \textit{noise-to-lane} paradigm can be analogous to the \textit{noise-to-image} process, 
where \textit{noise-to-image} generates the target image by removing the noise from noisy pixels and \textit{noise-to-lane} predicts the target lanes by removing the noise from the noisy lane anchors.
At the training stage, Gaussian noise is added to the starting point and the angle of the ground truth to obtain the noisy lane anchors. Then, the noisy lane anchors are sent to the RoI pooling~\cite{zheng2022clrnet} to crop the RoI features from the features of the encoder. Finally, 
the RoI features are fed into the decoder to predict the target lanes without noise. At the inference stage, DiffusionLane samples the random lane anchors from the Gaussian distribution and generates the target lanes by removing the noise in the random lane anchors.
However, the quality of the random lane anchors is poor compared to the learnable lane anchors, degrading the feature representation ability of the encoder. In order to solve this issue, we propose a hybrid decoding strategy, including the hybrid diffusion decoder and an auxiliary head. 
The former aims to refine the lane anchors with higher quality by combining the global-level and the local-level decoders, and the latter adopts the learnable lane anchors to provide extra positive supervision, which is only used in the training.

Experimental results demonstrate that our DiffusionLane achieves state-of-the-art results on four benchmarks, i.e., Carlane, Tusimple, CULane, and LLAMAS. For instance, DiffusionLane with MobileNetV4~\cite{qin2024mobilenetv4} gets 81.32\% F1 score on CULane, setting a new state-of-the-art result. 
Thanks to the random lane anchors and denoising diffusion paradigm, DiffusionLane shows a strong generalization ability and does not require re-training on the distribution-shift scenarios. For example, DiffusionLane achieves 86.23\%
accuracy on MuLane, a sub-dataset of the domain adaptation dataset Carlane, surpassing the CLRerNet~\cite{honda2024clrernet} by 1.21\% accuracy. The main contributions of this paper are as follows.

\begin{enumerate}
    \item We formulate the lane detection as the denoising diffusion process from the noisy lane anchors to the target lanes. To the best of our knowledge, we are the first to apply the diffusion model to the lane detection task.
    \item We propose a novel hybrid decoding strategy, including a hybrid diffusion decoder and an auxiliary head, to generate high-quality lane anchors and improve the feature representation ability of the encoder.
    \item Experimental results show that our DiffusionLane achieves the state-of-the-art performance on four benchmarks. Remarkably, DiffusionLane displays a strong generalization ability in the distribution-shift scenarios.
\end{enumerate}

\section{Related Work}

\subsection{Lane Detection}

Existing lane detection methods can be divided into three categories according to the representation of the lane: segmentation-based methods, anchor-based methods, and parameter-based methods.
Segmentation-based methods~\cite{pan2018spatial,zhang2023houghlanenet,zheng2021resa} regard the lane detection task as the segmentation task. SCNN~\cite{pan2018spatial} proposes a message-passing module to capture the spatial dependency. RESA~\cite{zheng2021resa} designs a real-time feature aggregation module to capture 
the global and local features while keeping real-time detection.
Anchor-based methods refine the predefined lane anchors to regress accurate lanes. UFLD~\cite{qin2020ultra} predicts the lanes by the novel row-wise anchor. Line-CNN~\cite{li2019line} adopts the dense lane anchors and the RoI pooling~\cite{ren2015faster} to achieve lane detection.
CLRNet~\cite{zheng2022clrnet} proposes learnable lane anchors and progressive lane refinement to detect lanes. Parameter-based methods~\cite{tabelini2021polylanenet,liu2021end,feng2022rethinking} cast the lane detection as a parametric modeling task.
PolyLaneNet~\cite{tabelini2021polylanenet} views a lane as a polynomial and predicts the parameters of the polynomial. LSTR~\cite{liu2021end} adopts the DETR-like architecture and realizes the end-to-end lane detection. Different from the existing methods, DiffusionLane reformulates the lane detection as a \textit{noise-to-lane} process. 

\subsection{Diffusion Model}

Diffusion model shows strong capabilities in the visual generation task~\cite{song2019generative,rombach2022high,zhang2023adding,blattmann2023stable}. Considering the strength of the diffusion model, several works transfer the diffusion model to the visual perception task.
DDP~\cite{ji2023ddp} concatenates the noisy feature map with the feature map outputted from the backbone and performs the semantic segmentation via the diffusion process. DiffusionTrack~\cite{xie2024diffusiontrack} views the object tracking task as the denosing task and develops a point set-based diffusion model to facilitate the object tracking. 
DiffusionDet~\cite{chen2023diffusiondet} predicts the bounding boxes from the noisy boxes. LSR-DM~\cite{ruiz2024lane} segments the lane graph with the diffusion model from the aerial imagery, not belonging to the traditional lane detection that detects lanes from an on-board camera. Our DiffusionLane is the first attempt to employ the diffusion denoising process in the lane detection task. 

The main difference between the previous visual diffusion models and DiffusionLane is the diffusion target. Existing methods add noise to the bounding box~\cite{chen2023diffusiondet} or the feature map~\cite{ji2023ddp}, while our method diffuses the starting point and the angle of the lane. Besides, we propose a hybrid decoding strategy to reduce
the influence of random lane anchors.

\section{Method}

In this section, we first introduce the preliminaries on the representation of the lane and the diffusion model. Then, we detail the model architecture, the training stage, and the inference stage. Finally, we describe the proposed hybrid decoding strategy.

\begin{figure}[t!]
    \centering
    \includegraphics[width=0.45\textwidth]{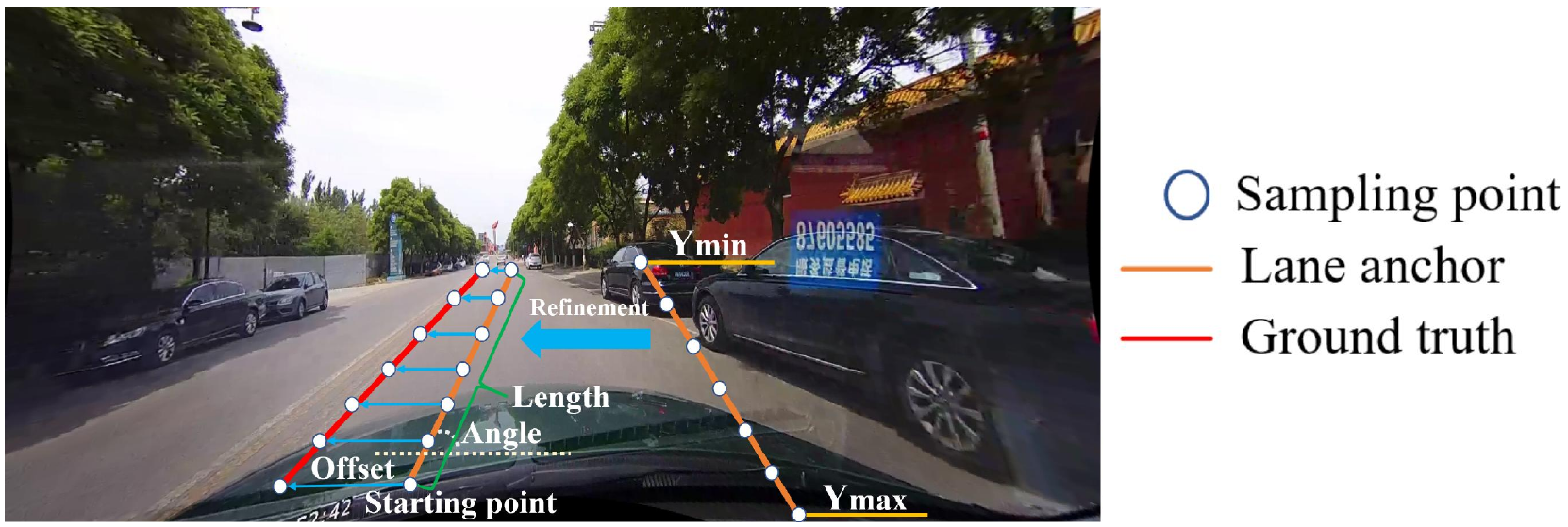}
        \caption{Illustration of lane representation. We take 7 sampling points as the example.}
    \label{fig:2}
    % \vspace{-0.3cm}
\end{figure}

\begin{figure*}[t!]
    \centering
    \includegraphics[width=0.8\textwidth]{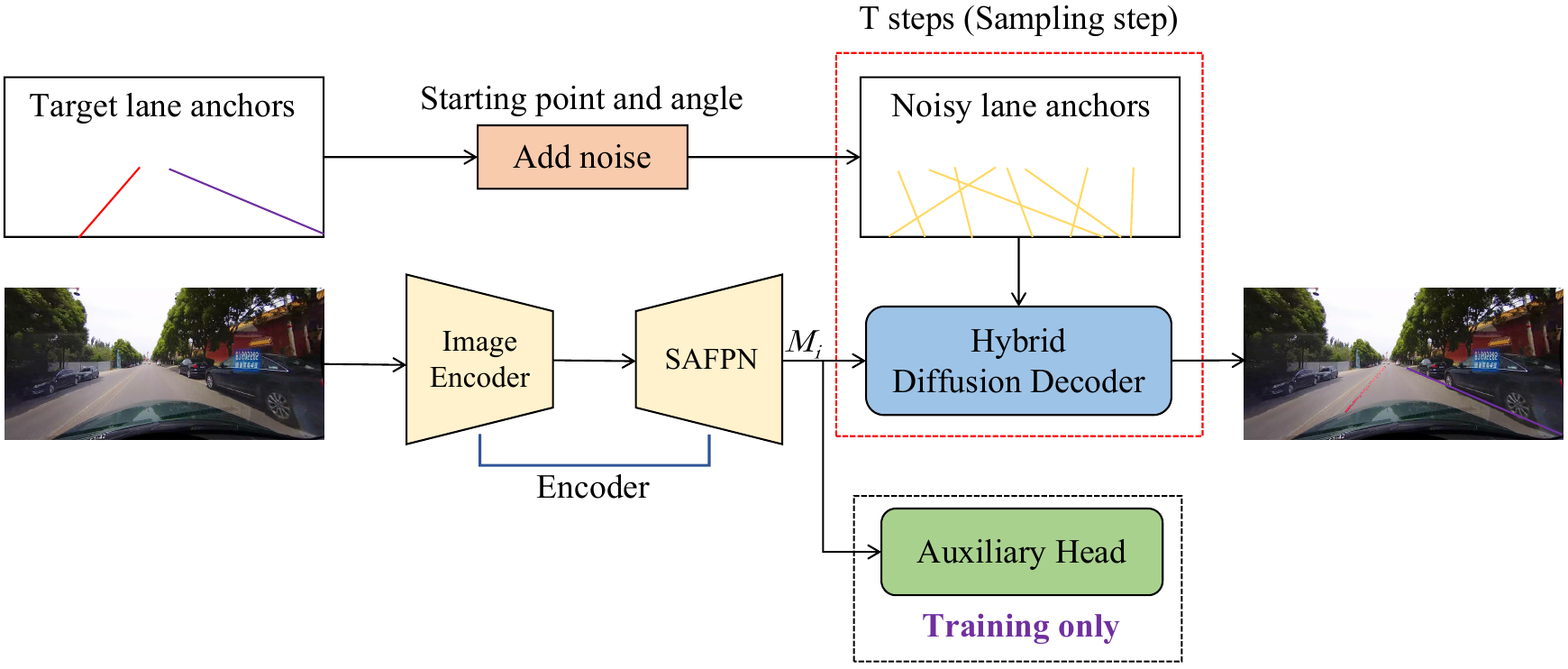}
        \caption{Training pipeline of DiffusionLane. The image encoder and SAFPN~\cite{xiao2023adnet} extract the multi-scale image features $M_i,i \in [0,2]$ and feed them into the hybrid diffusion decoder. The hybrid diffusion decoder, containing a stack of hybrid diffusion blocks, refines the noisy lanes to target lanes iteratively.
                Auxiliary head, only used in the training process, improves the feature representation of the encoder via learnable lane anchors.}
    \label{fig:3}
    % \vspace{-0.5cm}
\end{figure*}

% \section{Method}

% In this section, we first introduce the preliminaries on the representation of the lane and the diffusion model. Then, we detail the model architecture, the training stage, and the inference stage. Finally, we describe the proposed hybrid decoding strategy.

% \begin{figure}[t!]
%     \centering
%     \includegraphics[width=0.45\textwidth]{images/rep.pdf}
%         \caption{Illustration of lane representation. We take 7 sampling points as the example.}
%     \label{fig:2}
%     \vspace{-0.3cm}
% \end{figure}

% \begin{figure*}[t!]
%     \centering
%     \includegraphics[width=0.8\textwidth]{images/figure2.pdf}
%         \caption{Training pipeline of DiffusionLane, which consists of an encoder-decoder structure. The image encoder and SAFPN~\cite{xiao2023adnet} extract the image feature and feed it into the hybrid diffusion decoder. The hybrid diffusion decoder, containing a stack of hybrid diffusion blocks, refines the noisy lanes to target lanes iteratively.
%                 Auxiliary head, only used in the training process, improves the feature representation of the encoder via learnable lane anchors.}
%     \label{fig:3}
%     \vspace{-0.5cm}
% \end{figure*}

\subsection{Preliminaries}
\textbf{Lane representation}. Following~\cite{zheng2022clrnet}, a lane is represented as a sequence of of equal-spaced 2d points, i.e., $P=\{(x_{1},y_{1}),(x_{2},y_{2}),...,(x_{N},y_{N})\}$, where $N$ is the number of points. In this paper, we call the representation the \textit{lane anchor}. As shown in Figure~\ref{fig:2}, the $y$ coordinate is equally sampled along the image height, i.e., $y_{i} = \frac{Y_{max} - Y_{min}}{N-1} \times i$, 
here $Y_{max}$ and $Y_{min}$ denote the $y$ coordinates of the ending point and the starting point of the lane. In this paper, $Y_{max}$ is the image height, and $Y_{min}$ is set to a fixed value. Accordingly, the $x$ coordinates are one-to-one corresponding to the $y$ coordinates. DiffusionLane regresses the accurate lanes by refining the noisy lane anchors. The outputs of the model consist of four components: (1) the probabilities of background and foreground. (2) the starting point and the angle of the refined lanes.
(3) the length of the refined lanes. (4) $N$ offests, i.e., the horizontal distance between $N$ sampling points and their ground truth. We first predict the background and foreground probabilities of a lane anchor. Then, we obtain the starting point, angle, and length of a foreground lane anchor. Finally, we refine the foreground lane anchors by regressing $N$ offsets.

\textbf{Diffusion model}. Diffusion models~\cite{ho2020denoising,sohl2015deep,song2020denoising,song2019generative} are generative likelihood-based models, drawing inspiration from nonequilibrium thermodynamics~\cite{song2019generative,song2020improved}. Diffusion model defines two Markovian chains, i.e., a diffusion forward chain that adds noise to the image and a reverse chain that refines the noise back to the image.
Formally, given the distribution of an image $x_0 \sim q(x_0)$, the diffusion forward process at time $t$ is defined as $q(x_t | x_{t-1})$. During the diffusion forward process, the diffusion model gradually adds the Gaussian noise to the image according to the variance schedule $\beta_{1},...,\beta_{T}$, where $T$ denotes the total number of time steps. The diffusion forward process can be written as:

\begin{equation}
q(x_t|x_{t - 1}) = \mathcal{N}(x_t; \sqrt{1 - \beta_t} x_{t - 1}, \beta_t \mathbf{I})
\label{eq:1}
\end{equation}
Where $\mathbf{I}$ is the identity matrix. Benefiting from the Markovian chain, $x_t$ can be obtained by giving $x_0$:

\begin{equation}
    x_t = \sqrt{\bar{\alpha}_t} x_0 + (1 - \bar{\alpha}_t) \epsilon
    \label{eq:2}
\end{equation}
where $\epsilon \sim \mathcal{N}(0,\mathbf{I})$ represents the sampled Gaussian vector. $\bar{\alpha}_t = \prod_{s = 0}^t (1 - \beta_s)$. During the training process, the diffusion model predicts $x_0$ from $x_t$ at time $t$. In the inference, the model reverses the Gaussian noise $x_T$ back to $x_0$. 

% \begin{figure*}[t!]
%     \centering
%     \includegraphics[width=0.8\textwidth]{images/figure2.pdf}
%         \caption{Architecture of DiffusionLane. It consists of an encoder-decoder structure. The image encoder and SAFPN~\cite{xiao2023adnet} extract the image feature and feed it into the diffusion decoder. The diffusion decoder, containing a stack of diffusion layers, refines the noisy lanes to target lanes iteratively.
%                 Auxiliary head, only used in the training process, improves the feature representation of the model.}
%     \label{fig:3}
% \end{figure*}

\subsection{Diffusion Model for Lane Detection}
The overall architecture of DiffusionLane is shown in Figure~\ref{fig:3}, which is composed of the encoder (image encoder and SAFPN~\cite{xiao2023adnet}), the hybrid diffusion decoder, and the auxiliary head. The encoder takes the raw image as input and outputs the image features. The hybrid diffusion decoder receives the image features and runs $T$ times 
to obtain the predictions from the noisy lane anchors. Previous methods~\cite{blattmann2023stable,wu2024medsegdiff} requires multiple runs of the whole model to generate refined samples, while DiffusionLane only applies the 
decoder in multiple steps on the image features, significantly decreasing the computational burden. The auxiliary head is introduced to improve the feature representation of the encoder by adopting the learnable lane anchors.
The detailed description of the hybrid diffusion decoder and the auxiliary head is presented in the hybrid decoding section.

\textbf{Encoder}. As depicted in Figure~\ref{fig:3}, the encoder contains an image encoder and the SAFPN, an improved FPN with large kernel attention. The image encoder is responsible for extracting the high-level features of the input image. The extracted features are fed into SAFPN to generate multi-scale features. Single-scale feature is represented as $M_i,i \in [0,2]$.

\subsection{Training}
In the training stage, we first introduce the diffusion process, i.e., converting the ground truth to noisy lane anchors, and then train the model to reverse the diffusion process. We show 
the training pseudo-code of our method in Algorithm 1 of the supplementary materials.

\textbf{Ground truth padding}. In the existing lane detection benchmarks, the number of lanes varies across images, which is inconvenient for the training. In order to set the number of lanes across images to a fixed number $N_{train}$, we apply the $padding$ operation to the ground truth in each image. Specifically, we pad some extra lane anchors to the ground truth. Common padding strategies are repeating the ground truth and 
concatenating the random lane anchors. We adopt the concatenating random lane anchors since this strategy works best (see Table~\ref{tab:8}). 
% \clearpage

\textbf{Lane corruption}. We add the Gaussian noise to the padded ground truth. Existing methods~\cite{xie2024diffusiontrack,chen2023diffusiondet} add the Gaussian noise to the keypoints of the ground truth. 
However, adding the Gaussian noise to all points of a lane anchor leads to a heavy computational burden and hard optimization during the diffusion process. We add the 
Gaussian noise to the starting point and angle of a lane anchor since the starting point and the angle determine the coarse location of a lane anchor. As presented in Figure~\ref{fig:3}, given a ground true lane, 
% we first sample a Gaussian vector $\epsilon \in \text{R}^{1 \times 3}$ and add it to the starting point and the angle of the ground truth $b_0 \in \text{R}^{1 \times 3}$ by
we first extend the $y$ coordinates of the starting point and the ending point to $Y_{min}$ and $Y_{max}$ to obtain the target lane anchor following~\cite{zheng2022clrnet}.
Then, we sample a Gaussian vector $\epsilon \in \text{R}^{1 \times 3}$ and add it to the starting point coordinate $X_0,Y_0$ and the angle $\theta_0$ of the target lane anchor by
% here $x_0[:4]$ denotes the x,y coordinates of starting point, angle, and length. Then, we sample a Gaussian vector $\epsilon \in \text{R}^{1 \times 3}$ and add $\epsilon$ to $x_0[:3]$ by 
\begin{equation}
    [X_t,Y_t,\theta_t] = \sqrt{\bar{\alpha}_t} [X_0,Y_0,\theta_0] + (1 - \bar{\alpha}_t) \epsilon
    \label{eq:3}
\end{equation}
% where $b_0$ denotes the $x,y$ coordinates of the starting point and the angle.
Noise scale is controlled by $\bar{\alpha}_t$, which is adjusted by the cosine schedule at time step $t$. Finally, we get the noisy lane anchor according to $[X_t,Y_t,\theta_t]$. 

\textbf{Training losses}. We adopt focal loss~\cite{lin2017focal} as the classification loss, segmentation loss~\cite{qin2020ultra} as the auxiliary loss, smooth L1 loss, angle loss~\cite{su2024gsenet}, and Line-IoU loss~\cite{zheng2022clrnet} as the regression loss. 
The weight of angle loss is 0.02, and the weights of other losses are the same as those in CLRNet. Losses are calculated between $N_{train}$ refined lanes outputted by the hybrid diffusion decoder and the ground truths.
SimOTA algorithm~\cite{zheng2022clrnet} is utilized to assign multiple predictions to each ground truth.

\subsection{Inference}
The inference process of DiffusionLane reverses the noisy lane anchors to target lanes. Starting from the noisy lane anchors sampled from the Gaussian distribution, DiffusionLane refines the noisy lanes progressively.
Pseudo-code of the inference stage is offered in Algorithm 2 of the supplementary materials.

\textbf{Sampling step}. The hybrid diffusion decoder runs $T$ times to perform lane anchor refinement, and each refinement is called a sampling step. Specifically, the random lane anchors at the first sampling step or the refined lane anchors from the last sampling step are fed to the hybrid diffusion decoder to obtain further refined lane anchors. Then, the refined lane anchors are sent to the next sampling for iterative refinement via DDIM~\cite{song2020denoising}.
As done in lane corruption in the training, we sample the starting points and angles from the Gaussian distribution to construct random lane anchors at the first sampling step.

\textbf{Lane anchors resampling}. After the refinement in each sampling step, the lane anchors can be divided into two categories: foreground and background. Common practice is to filter the background via setting the confidence threshold. However, the number of foreground lane anchors varies across the image, leading to a conflict with ground truth padding in the training.
We adopt the lane anchor resampling strategy to solve this conflict.
% we find that directly sending the foreground lane anchors to the next sampling step leads to performance degradation. We attribute the reason to feature distribution shift between the lane corruption in the training and the foreground lane anchors. We adopt the lane anchor resampling strategy to align the feature distribution between the inference and the training. 
Specifically, we first filter the background lane anchors and then 
concatenate the foreground lane anchors with the random lane anchors sampled from the Gaussian distribution, which can ensure the feature distribution alignment between the inference and the training. The concatenated lane anchors are sent to the next sampling step. After finishing all sampling steps, we adopt NMS algorithm~\cite{zheng2022clrnet} to remove the duplicate predictions in the foreground lane anchors.

\begin{figure}[t!]
    \centering
    \includegraphics[width=0.47\textwidth]{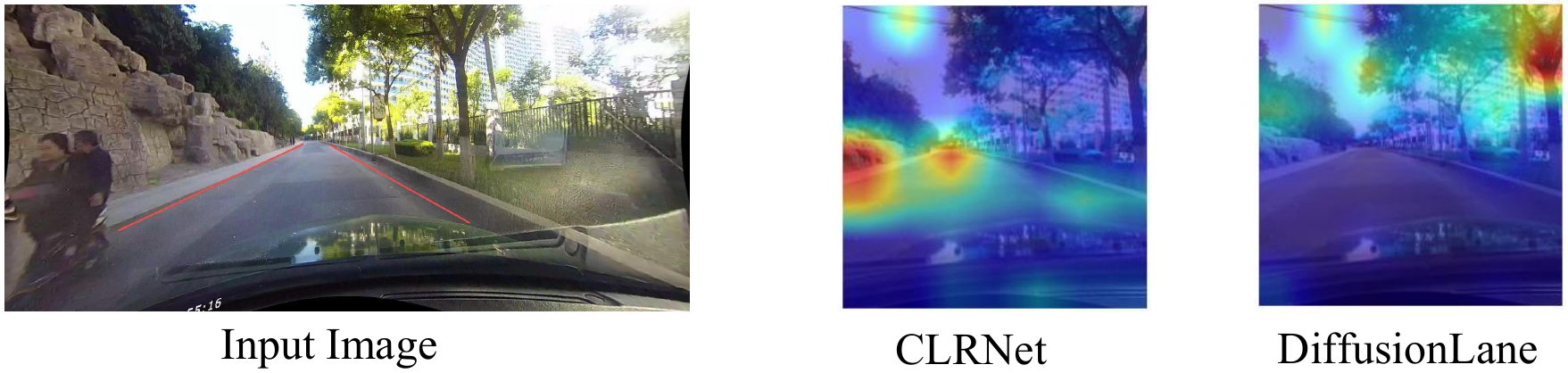}
        \caption{Visualization of the attention map of the encoder.}
    \label{fig:5}
    % \vspace{-0.5cm}
\end{figure}

\subsection{Hybrid Decoding}
Recent studies~\cite{xiao2023adnet,wang2024polar} point out that the quality of the lane anchors has an impact on the feature representation of the encoder. DiffusionLane samples lane anchors from the Gaussian distribution and the quality of initial lane anchors is not as good as 
the learnable lane anchors used in previous methods~\cite{zheng2022clrnet,honda2024clrernet}. As shown in Figure~\ref{fig:5}, the encoder of CLRNet can discriminate the features of the lane better than DiffusionLane. To alleviate the poor representation caused by random lane anchors, we propose a novel hybrid decoding strategy, including a novel hybrid diffusion decoder and an auxiliary head with learnable lane anchors.

\begin{figure}[t!]
    \centering
    \includegraphics[width=0.4\textwidth]{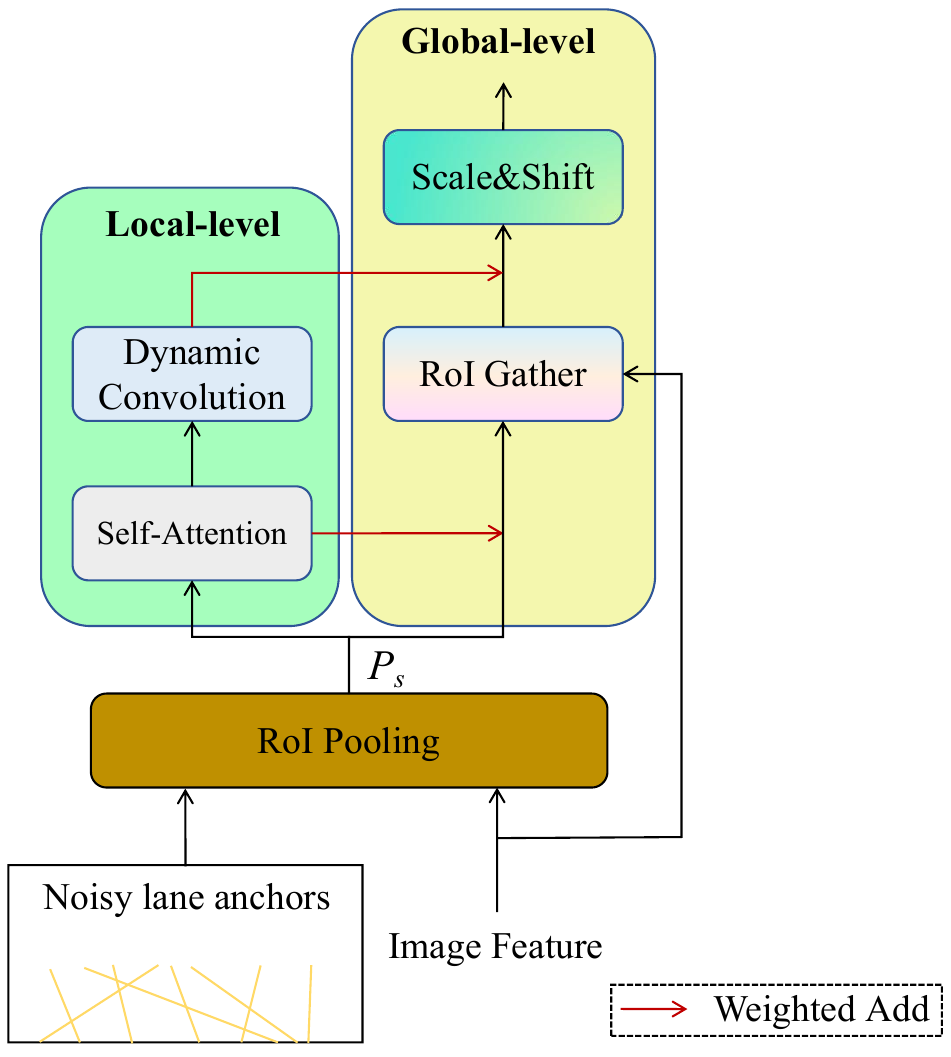}
        \caption{Architecture of a hybrid diffusion block.}
    \label{fig:4}
    % \vspace{-0.5cm}
\end{figure}

\textbf{Hybrid diffusion decoder}. In general, a lane detector is equipped with a single decoder, while we argue that a lane detector can utilize multiple decoders. Although multiple decoders bring an extra computational burden, multiple decoders can integrate the strengths of each decoder and complement each other to generate high-quality lane anchors. 
Based on this point, we propose a novel hybrid diffusion decoder, which combines two decoders at both global and local levels. To be specific, the hybrid diffusion decoder consists of a stack of hybrid diffusion blocks and each block corresponds to a scale image feature $M_{i}$.
The architecture of a single hybrid diffusion block is shown in Figure~\ref{fig:4}.
Each hybrid diffusion block takes the lane anchors from the previous block and image feature $M_i$ as inputs. The first diffusion block receives the noisy lane anchors.
We define the set of RoI features of lane anchors $P_s={\{p_{i}\}}_{i=0}^{N_{train}}$ generated by RoI pooling~\cite{tabelini2021keep}.

For the global-level decoder, we adopt the RoIGather module~\cite{zheng2022clrnet} to aggregate the global feature into $p_{i}$. Since the noise scale has an important impact on the performance of the diffusion model~\cite{chen2023generalist}, we apply the scale and shift operation~\cite{chen2023diffusiondet} to $p_{i}$.

For the local-level decoder, we utilize a self-attention module to enhance $P_s$ and then adopt the dynamic convolution~\cite{sun2021sparse} to fuse the local features into $p_i$. Dynamic convolution enhances the local features by facilitating the interaction between $p_i$ and other roi features. Specifically, we take the $P_s$ as the convolution kernels. Thereafter,
these kernels are applied to $p_{i}$ via convolution layers. To take full advantage of the potential of each decoder, we view the global-level decoder as the main decoder and the local-level decoder as the auxiliary decoder. The local-level decoder injects the implicit information into the input and output features of RoIGather. In particular, we integrate the output features of 
self-attention module and dynamic convolution into the input and output features of RoIGather through a simple add operation with learnable weights. The output features of the main decoder are used to generate refined lane anchors.

\textbf{Auxiliary head}. The poor lane anchors cause sparse supervision on the encoder due to the fewer positive samples, damaging the feature representation of the encoder. To alleviate this issue, we introduce an auxiliary head with learnable lane anchors, e.g., the head of CLRNet or CLRerNet, to enrich the supervision on the encoder. As shown in Figure~\ref{fig:3}, we send the multi-scale features $M_{i}$ into the auxiliary head to get the predictions $Q_{i}$. 
The auxiliary head computes the supervised targets for positive and negative samples in $Q_{i}$. The training losses of the auxiliary head are the same as the original head, e.g., CLRNet head or CLRerNet head. The auxiliary head is only used in the training process.

% \begin{figure}[t!]
%     \centering
%     \includegraphics[width=0.5\textwidth]{images/fig4.pdf}
%         \caption{Visualization of the attention map of the image encoder.}
%     \label{fig:5}
%     % \vspace{-0.5cm}
% \end{figure}

\section{Experiments}
\subsection{Datasets}
% We conduct the experiments on four lane detection benchmarks: Carlane~\cite{stuhr2022carlane}, LLAMAS~\cite{behrendt2019unsupervised}, Tusimple~\cite{shirke2019lane}, and CULane~\cite{pan2018spatial}.

We conduct the experiments on four lane detection benchmarks: CULane~\cite{pan2018spatial}, LLAMAS~\cite{behrendt2019unsupervised}, Tusimple~\cite{shirke2019lane}, and Carlane~\cite{stuhr2022carlane}.

\textbf{Carlane} is a dataset for domain adaptative lane detection task, containing three sub-datasets: TuLane, MoLane, and MuLane. Each sub-dataset comprises over 20K training images. Image size is 1280$\times$720.

\textbf{LLAMAS} is a large-scale lane detection datasets with over 100K images. All lanes are annotated with accurate maps. Image size is 1280$\times$717.

\textbf{Tusimple} is a dataset for highway scenario, consisting of 3626 images for training and 2782 images for testing. All images have 1280$\times$720 pixels.

\textbf{CULane} is a widely used dataset for lane detection, which is composed of 88.9K training images, 9.7K images for validation, and 34.7K testing images. Image size is 1640$\times$590.

% The evaluation metrics are offered in the supplementary materials.

% \textbf{Carlane} is a dataset for domain adaptative lane detection task, containing three sub-datasets: TuLane, MoLane, and MuLane. Each sub-dataset is composed of over 20K training images. The source domain of TuLane adopts 24000 labeled simulated images for training and the target domain derives from Tusimple dataset. The source domain of MoLane has 80000 labeled simulated images as the training set and target domain is collected from the real scenes with 43843 unlabeled images.
% The MuLane dataset mixes the TuLane and MoLane dataset, which uses 48000 labeled simulated images as the training set and the target domain combines the target domain of MuLane and TuLane. Image size is 1280$\times$720.

% \subsection{Implementation Details}
% We introduce the implementation details and the evaluation metrics in the supplementary materials.

\subsection{Evaluation Metrics}
We utilize F1 score to measure the performance for CULane and LLAMAS datasets. For Tusimple and Carlane datasets, we adopt accuracy, FP, and FN to evaluate the model performance. More details of the evaluation metrics are provided in the supplementary materials.
% % For CULane and LLAMAS dataset, we utilize F1 score to measure the performance: $F_1 = \frac{2 \times Precision \times Recall}{Precision + Recall}$, where $Precision=\frac{TP}{TP+FP}$ and $Recall=\frac{TP}{TP+FN}$. $TP$, $FP$, and $FN$ represent the true positive rate, the false positive rate, and the false negative rate, respectively.

% % For Tusimple and Carlane datasets, we adopt accuracy, FP, and FN to evaluate the model performance. Accuracy is defined as $Accuracy = \frac{\sum_{clip} C_{clip}}{\sum_{clip} S_{clip}}$, $C_{clip}$ denotes the number of accurately predicted lane points and $S_{clip}$ represents the total number of lane points of a clip. A lane point is treated as a correct point if its distance is smaller than the threshold $t_{pc} = \frac{20}{cos(a_{yl})}$, here $a_{ul}$
% % is the angle of the corresponding ground truth.

\subsection{Implementation Details}
% % We select the ResNet~\cite{he2016deep}, DLA~\cite{yu2018deep}, and MobileNetV4-Hybrid-M~\cite{qin2024mobilenetv4} as the image encoder and all image encoders are initialized with pretrained weights on ImageNet1K~\cite{deng2009imagenet}. We train DiffusionLane using AdamW optimizer with learning rate 0.0003. Cosine schedule is adopted to adjust the learning rate. All models are trained on a single 3090 GPU with 24 GB memory and batch size is 20. 
% % The training epoches are set to 70,25,20,20 for Tusimple, CULane, LLAMAS, and Carlane datasets, respectively. The total time steps $T$ is set to 2. The number of lanes $N_{train}$ in the ground truth padding is 800. We choose CLRNet head as the auxiliary head. The confidence threshold to distinguish the foreground and background is 0.4. The initial noise scale is set to 2. The number of hybrid diffusion block is 3.
% % We set $Y_{min}$ to 160, 270, 300, and 160 for Tusimple, CULane, LLAMAS, and Carlane.  The number of lanes $N_{train}$ in the ground truth padding is 800. 

The total time steps $T$ is set to 2. We choose CLRNet head as the auxiliary head. The confidence threshold to distinguish the foreground and background is 0.4. The initial noise scale is set to 2. The number of hybrid diffusion block is 3. More implementation details are in the supplementary materials.

\begin{table}[t]
    \centering
    \caption{Performance on MoLane and TuLane. Acc denotes the Accuracy and * represents Source-only.}
    \Huge
    \resizebox{0.5\textwidth}{!}{
    % \fontsize{size}{skip}
    \begin{tabular}{l c c c c c c}
    \toprule
    \multirow{2}{*}{Methods} & \multicolumn{3}{c}{MoLane} & \multicolumn{3}{c}{TuLane} \\
    \cmidrule(lr){2-4} \cmidrule(l){5-7}
    % \cline{2-4} \cline{5-7}
    & Acc(\%)$\uparrow$ & FP(\%)$\downarrow$ & FN(\%)$\downarrow$ & Acc(\%)$\uparrow$ & FP(\%)$\downarrow$ & FN(\%)$\downarrow$ \\
    \midrule
    UFLD*~\cite{qin2020ultra} & 88.15 & 34.35 & 28.45 & 87.43 & 34.21 & 23.48 \\
    BezierLaneNet*~\cite{feng2022rethinking} & 86.78 & 30.15 & 28.10 & 86.10 & 32.99 & 24.21 \\ 
    CLRNet*~\cite{zheng2022clrnet} & 89.14 & 27.00 & 26.96 & 87.92 & 32.25 & 20.28 \\ 
    % Lane2Seq (segmentation)*~\cite{zhou2024lane2seq} & 88.98 & 27.12 & 27.85 & 87.28 & 33.95 & 22.60 \\ 
    DACCA*~\cite{zhou2024unsupervised} & 86.15 & 38.85 & 29.50 & 85.93 & 34.99 & 24.22 \\
    CLRerNet*~\cite{honda2024clrernet} & 88.54 & 32.26 & 28.33 & 87.35 & 34.56 & 23.44 \\
    DANN~\cite{ganin2016domain} & 85.25 & 39.07 & 36.18 & 88.74& 32.71 & 21.64 \\
    % \rowcolor{gray!25}
    DiffusionLane* & \bf 91.28 & \bf 24.59 & \bf 19.32 & \bf 91.00 & \bf 26.11 & \bf 15.77 \\
    % DANN & 89.58 & 28.75 & 23.24 & 91.06 & 30.17 & 18.54 \
    % ADDA & 91.76 & 21.04 & 13.06 & 91.39 & 28.76 & 16.63 \
    % SGADA & 92.59 & 18.32 & 9.88 & 92.04 & 28.18 & 15.99 \
    % SGPCS& 92.82 & 17.10 & 8.77 & 93.29 & 25.68 & 12.73 \
    % Proposed & \textbf{94.5} & \textbf{15.4} & \textbf{5.95} & \textbf{93.57} & \textbf{23.23} & \textbf{9.8} \
    % Target-only & 96.92 & 0.94 & 0.03 & 94.43 & 20.74 & 7.20 \
    \bottomrule
    \end{tabular}
    }
    \label{tab:1}
\end{table}

\begin{table}[t!]
    \centering
    \small
    \caption{Performance comparison of different models on MuLane.}
    \resizebox{\columnwidth}{!}{
        \begin{tabular}{lccc}
            \hline
            Methods& Acc(\%)$\uparrow$ & FP(\%)$\downarrow$ & FN(\%)$\downarrow$ \\
            \hline
            UFLD*~\cite{qin2020ultra} & 79.61 & 44.78 & 33.36 \\
            BezierLaneNet*~\cite{feng2022rethinking} & 77.23 & 46.26 & 35.11 \\
            CLRNet*~\cite{zheng2022clrnet} & 84.44 & 39.62 & 29.82 \\
            CLRerNet*~\cite{honda2024clrernet} & 85.01 & 35.52 & 27.10 \\
            % Lane2Seq (segmentation)*~\cite{zhou2024lane2seq} & 83.15 & 41.14 & 29.58 \\
            DACCA*~\cite{zhou2024unsupervised} & 83.28 & 47.17 & 55.21 \\
            DANN~\cite{ganin2016domain} & 84.01 & 38.31 & 36.30 \\
            % \rowcolor{gray!25}
            DiffusionLane* & \bf 86.23 & \bf 27.85 & \bf 25.31 \\
            \hline
        \end{tabular}
    }
    \label{tab:2}
    % \vspace{-0.3cm}
\end{table}

\subsection{Comparison with the State-of-the-art Methods}
\textbf{Performance on Carlane}. We first compare the domain adaptation ability of different lane detectors. Source-only means model is only accessible to the source domain in the training process.
We adopt ResNet18 as the image encoder for all models.
Results are shown in Table~\ref{tab:1} and~\ref{tab:2}. We can see that DiffusionLane achieves the best performance among the existing methods. Compared with CLRerNet, DiffusionLane
gains 2.74\% (91.28\% vs. 88.54\%), 3.65\% (91.00\% vs. 87.35\%), and 1.22\% (86.23\% vs. 85.01\%) accuracy improvements on MoLane, TuLane, and MuLane. Compared with the domain adaptation method DANN with UFLD,
DiffusionLane obtains the better performance on three datasets. The results manifest the strong generalization ability of DiffusionLane in the domain shift scenarios. We attribute the reason to that the lane anchors of DiffusionLane
are sampled from Gaussian distribution, which are domain agnostic. However, the lane anchors in existing methods like CLRNet are learned from the source domain, requiring the data distribution is similar between 
the target domain and the source domain.

\begin{table}[t!]
    \centering
    % \small
    \caption{Performance comparison of different models on LLAMAS.}
    \resizebox{\columnwidth}{!}{
        \begin{tabular}{lc c}
            \hline
            Methods& Image encoder &F1(\%)$\uparrow$ \\
            \hline
            PolyLaneNet~\cite{tabelini2021polylanenet} & EfficientNet-B0~\cite{tan2019efficientnet} & 90.20 \\
            BezierLaneNet~\cite{feng2022rethinking} & ResNet34 & 96.11 \\
            LaneATT~\cite{tabelini2021keep} & ResNet34 & 94.96 \\
            LaneATT~\cite{tabelini2021keep} & ResNet122 & 95.17 \\
            LaneAF~\cite{abualsaud2021laneaf} & DLA34 & 96.90 \\
            CLRNet~\cite{zheng2022clrnet} & ResNet18 & 96.96 \\
            CLRNet~\cite{zheng2022clrnet} & DLA34 & 97.16 \\
            Lane2Seq (segmentation)~\cite{zhou2024lane2seq} & ViT-Base~\cite{dosovitskiy2020image} & 97.42 \\
            \hline
            % \rowcolor{gray!25}
            DiffusionLane & ResNet18 & 97.27 \\
            % \rowcolor{gray!25}
            DiffusionLane & ResNet34 & 97.36 \\
            % \rowcolor{gray!25}
            DiffusionLane & ResNet101 & \bf 97.59 \\
            \hline
        \end{tabular}
    }
    \label{tab:3}
\end{table}

\textbf{Performance on LLAMAS}.
As shown in Table~\ref{tab:3}, DiffusionLane increases the F1 score from 97.42\% to 97.59\% compared to 
the previous state-of-the-art method Lane2Seq. DiffusionLane with ResNet18 achieves 0.31\% (97.27\% vs. 96.90\%) F1 score improvement compared to CLRNet with ResNet18.
The results show the effectiveness of DiffusionLane in multi-lane scenarios (the number of lanes $>$ 4).

\begin{table}[t!]
    \centering
    % \small
    \caption{Performance comparison of different models on Tusimple.}
    \resizebox{\columnwidth}{!}{
        \begin{tabular}{lc c  c c}
            \hline
            Methods& Image encoder &Acc(\%)$\uparrow$ & FP(\%)$\downarrow$ & FN(\%)$\downarrow$ \\
            \hline
            SCNN~\cite{pan2018spatial} & VGG16 & 96.53 & 6.17 & \bf 1.80 \\
            %RESA~\cite{zheng2021resa} & ResNet34 & 96.82 & 3.63 & 2.48 \\
            UFLD~\cite{qin2020ultra} & ResNet34 & 95.86 & 18.91 & 3.75 \\
            GANet~\cite{wang2022keypoint} & ResNet34 & 95.87 & 1.99 & 2.64 \\
            LaneATT~\cite{tabelini2021keep} & ResNet122 & 96.10 & 5.64 & 2.17 \\
            CondLaneNet~\cite{liu2021condlanenet} & ResNet34 & 95.37 & 2.20 & 3.82 \\
            CondLaneNet~\cite{liu2021condlanenet} & ResNet101 & 96.54 & 2.01 & 3.50 \\
            CLRNet~\cite{zheng2022clrnet} & ResNet34 & 96.87 & 2.27 & 2.08 \\
            CLRNet~\cite{zheng2022clrnet} & ResNet101 & 96.83 & 2.37 & 2.38 \\
            GSENet~\cite{zheng2022clrnet} & ResNet34 & 96.88 & 2.04 & - \\
            Lane2Seq (segmentation)~\cite{zhou2024lane2seq} & ViT-Base & 96.85 & 2.01 & 2.03 \\
            \hline
            % \rowcolor{gray!25}
            DiffusionLane & ResNet18 & 96.84 & 2.15 & 2.10 \\ 
            % \rowcolor{gray!25}
            DiffusionLane & ResNet34 & \bf 96.89 & 2.03 & 2.03 \\ 
            % \rowcolor{gray!25}
            DiffusionLane & ResNet101 & 96.86 & \bf 1.97 & 2.06 \\ 
            % BezierLaneNet~\cite{feng2022rethinking} & ResNet34 & 96.11 \\
            % LaneATT~\cite{tabelini2021keep} & ResNet34 & 94.96 \\
            % LaneATT~\cite{tabelini2021keep} & ResNet122 & 95.17 \\
            % LaneAF~\cite{abualsaud2021laneaf} & DLA34 & 96.90 \\
            % CLRNet~\cite{zheng2022clrnet} & ResNet18 & 96.96 \\
            % CLRNet~\cite{zheng2022clrnet} & DLA34 & 97.16 \\
            % Lane2Seq (segmentation)~\cite{zhou2024lane2seq} & ViT-Base~\cite{dosovitskiy2020image} & 97.42 \\
            % \hline
            % DiffusionLane & ResNet18 & \bf 97.27 \\
            % DiffusionLane & ResNet34 & \bf 97.36 \\
            % \rowcolor{gray!25}
            % DiffusionLane & ResNet101 & \bf 97.59 \\
            \hline
        \end{tabular}
    }
    \label{tab:4}
    % \vspace{-0.3cm}
\end{table}

\textbf{Performance on Tusimple}. We present the results on Tusimple dataset in Table~\ref{tab:4}. Since the data scale of Tusimple is small, the performance gap between different methods is small.
DiffusionLane with ResNet34 achieves the best accuracy of 96.89\%. DiffusionLane with ResNet101 decreases FP from 1.99\% to 1.97\%, establishing a new state-of-the-art result for this indicator.

% \begin{table}[t!]
%     \centering
%     \Large
%     \caption{Effectiveness of each component in DiffusionLane. CLRNet are adopted as the baseline.}
%     \resizebox{\columnwidth}{!}{
%         \begin{tabular}{cccccc}
%             \hline
%             Random lane anchors & Diffusion paradigm & Hybrid diffusion decoder & Auxiliary head & F1(\%) \\
%             \hline
%             & & & & 79.96 \\
%             \hline
%             \checkmark & & & & 74.74 \\
%             \checkmark & \checkmark & & & 78.38 \\
%             \checkmark & \checkmark & \checkmark& & 79.46 \\
%             \checkmark & \checkmark & \checkmark& \checkmark & 80.24 \\
%             % \checkmark & \checkmark & \checkmark& \checkmark & 80.24 \\
%             % & & & & 55.23 & 79.58 & 62.21 \
%             % \checkmark & & & & 55.61 & 79.81 & 62.76 \
%             % \checkmark & \checkmark & & & 55.61 & 80.03 & 63.09 \
%             % \checkmark & \checkmark & \checkmark & & 55.87 & 80.31 & \textbf{63.55} \
%             % \checkmark & \checkmark & \checkmark & \checkmark & 55.93 & \textbf{80.42} & 63.51 \
%             \hline
%             \end{tabular}
%     }
%     \label{tab:6}
% \end{table}

\begin{table*}[!t]
	\centering
	\caption{Comparison of F1 score on CULane testing set. We only report the false positives for “Cross” category.}
	\tabcolsep=0.5cm
	\Huge
	\resizebox{\textwidth}{!}{
	\begin{tabular}{cccccccccccc}
	  \toprule
	  Methods& Image encoder& Normal$\uparrow$ & Crowded$\uparrow$& Dazzle$\uparrow$& Shadow$\uparrow$& No line$\uparrow$& Arrow$\uparrow$& Curve$\uparrow$& Night$\uparrow$& Cross$\downarrow$& Total$\uparrow$ \\
	  \midrule
	  SCNN~\cite{pan2018spatial}&ResNet50& 90.60& 69.70& 58.50& 66.90& 43.40& 84.10& 64.40& 66.10& 1900& 71.60 \\
	  RESA~\cite{zheng2021resa}&ResNet50& 92.10& 73.10& 69.20& 72.80& 47.70& 88.30& 70.30& 69.90& 1503& 75.3 \\
	  AtrousFormer~\cite{yang2023lane}&ResNet34& 92.83& 75.96& 69.48& 77.86& 50.15& 88.66& 71.14& 73.74& 1054& 78.08\\
	  % LaneAF~\cite{yang2023lane}&DLA34& 91.80& 75.61& 71.78& 79.12& 51.38& 86.88& 72.70& 73.03& 1360& 77.41 \\
	%   \rowcolor{gray!25}
	%   Lane2Seq (segmentation)&ViT-Base& \bf 93.39& \bf 77.27& \bf 73.45& \bf 79.69& \bf 53.91& \bf 90.53& \bf 73.37& \bf 74.96& 1129& \bf 79.64& 15.19 \\
	  LaneATT~\cite{tabelini2021keep}&ResNet122& 91.74& 76.16& 69.47& 76.31& 50.46& 86.29& 64.05& 70.81& 1264& 77.02 \\
	  O2SFormer~\cite{zhou2023end}&ResNet50& 93.09& 76.57& 72.25& 76.56& 52.80& 89.50& 69.60& 73.85& 3118& 77.83 \\
	  UFLD~\cite{qin2020ultra}&ResNet34& 90.70& 70.20& 59.50& 69.30& 44.40& 85.70& 69.50& 66.70& 2037& 72.30 \\
	  CondLaneNet~\cite{liu2021condlanenet}&ResNet34& 93.38& 77.14& 71.17& 79.93& 51.85& 89.89& 73.88& 73.92& 1387& 78.74 \\
	  ADNet~\cite{xiao2023adnet}&ResNet34& 92.90& 77.45& 71.71& 79.11& 52.89& 89.90& 70.64& 74.78& 1499& 78.94 \\
	%   \rowcolor{gray!25}
	%   Lane2Seq (anchor)&ViT-Base& 93.11& 77.43& 73.25& 79.46& 53.74& 90.02& 72.44& \bf 75.12& \bf 1173& 79.27& 15.19 \\
	%  LSTR~\cite{liu2021end}&ResNet18& -& -& -& -& -& -& -& -& -& 64.00 \\
	  BezierLaneNet~\cite{feng2022rethinking}&ResNet18& 90.22& 71.55& 62.49& 70.91& 45.30& 84.09& 58.98& 68.70& 996& 73.67 \\
	  BSNet~\cite{chen2023bsnet}&ResNet34& 93.75& 78.01& 76.65& 79.55& 54.69& 90.72& 73.99& 75.28& 1445& 79.89 \\
	  Eigenlanes~\cite{jin2022eigenlanes}&ResNet50& 91.70& 76.00& 69.80& 74.10& 52.20& 87.70& 62.90& 71.80& 1509& 77.20 \\
	  Laneformer~\cite{han2022laneformer}&ResNet50&91.77& 75.74& 70.17& 75.75& 48.73& 87.65& 66.33& 71.04& {\bf 19}& 77.06 \\
      CLRNet~\cite{zheng2022clrnet}&ResNet34& 93.49& 78.06& 74.57& 79.92& 54.01& 90.59& 72.77& 75.02& 1216& 79.73\\
      Lane2Seq (segmentation)~\cite{zhou2024lane2seq}&ViT-Base&93.39& 77.27& 73.45& 79.69& 53.91& 90.53& 73.37& 74.96& 1129& 79.64 \\
      CLRerNet~\cite{han2022laneformer}&ResNet34&93.93& 79.51& 73.88& 83.16& 55.55& 90.87& 74.45& 76.02& 1088& 80.76 \\
      CLRerNet~\cite{han2022laneformer}&DLA34&94.02& \bf 80.20 & 74.41 & 83.71& 56.27 & 90.39& 74.67& 76.53& 1161& 81.12 \\
      \midrule
      DiffusionLane & ResNet34 & 93.82& 78.65& 74.39& 82.18& 54.86 & \bf 90.88& 73.77& 75.79& 1119& 80.24\\
      DiffusionLane & ResNet50 & 93.91& 79.25& 74.66& 82.57& 55.32 & 90.69& 73.61& 76.06& 1054& 80.68\\
      DiffusionLane & DLA34 & 94.06 & 79.94 & \bf 74.78& 83.75& 56.24 & 90.45& 74.80 & 76.69& 1233& 81.18\\
    %   \rowcolor{gray!25}
      DiffusionLane & MobileNetV4-Hybrid-M & \bf 94.15 & 79.99 & 74.57& \bf 83.80& \bf 56.45 & 90.41& \bf 75.02 & \bf 76.87& 1133& \bf 81.32\\
	%   \rowcolor{gray!25}
	%   Lane2Seq (parameter)&ViT-Base& 93.03& 76.42& 72.17& 78.32& 52.89& 89.67& 72.67& 73.98& 1319& 78.39& 15.19 \\
	  \bottomrule
	\end{tabular}}
	\label{tab:5}
  \end{table*}

\textbf{Performance on CULane}. The benchmark results on CULane dataset are presented in Table~\ref{tab:5}. It is worth noting that DiffusionLane with MobileNetV4-Hybrid-M
reaches a new state-of-the-art F1 score of 81.32\%. DiffusionLane with DLA34 achieves the better performance than the previous state-of-the-art method CLRerNet (81.18\% vs. 81.12\%), especially in difficult 
scenarios such as Dazzle (74.78\% vs. 74.41\%) and Night (76.79\% vs. 76.53\%). Compared with CLRNet, DiffusionLane with ResNet34 surpasses it by 0.49\% (80.24\% vs. 79.73\%) F1 score improvement. The results indicate that 
the random lane anchors can achieve competitive even better performance than the learnable lane anchors with the proposed diffusion paradigm and the hybrid decoding strategy. Besides,
compared with the parameter-based method BSNet and Lane2Seq with segmentation format, DiffusionLane also has the performance advantages.

\begin{figure}[t!]
    \centering
    \includegraphics[width=0.47\textwidth]{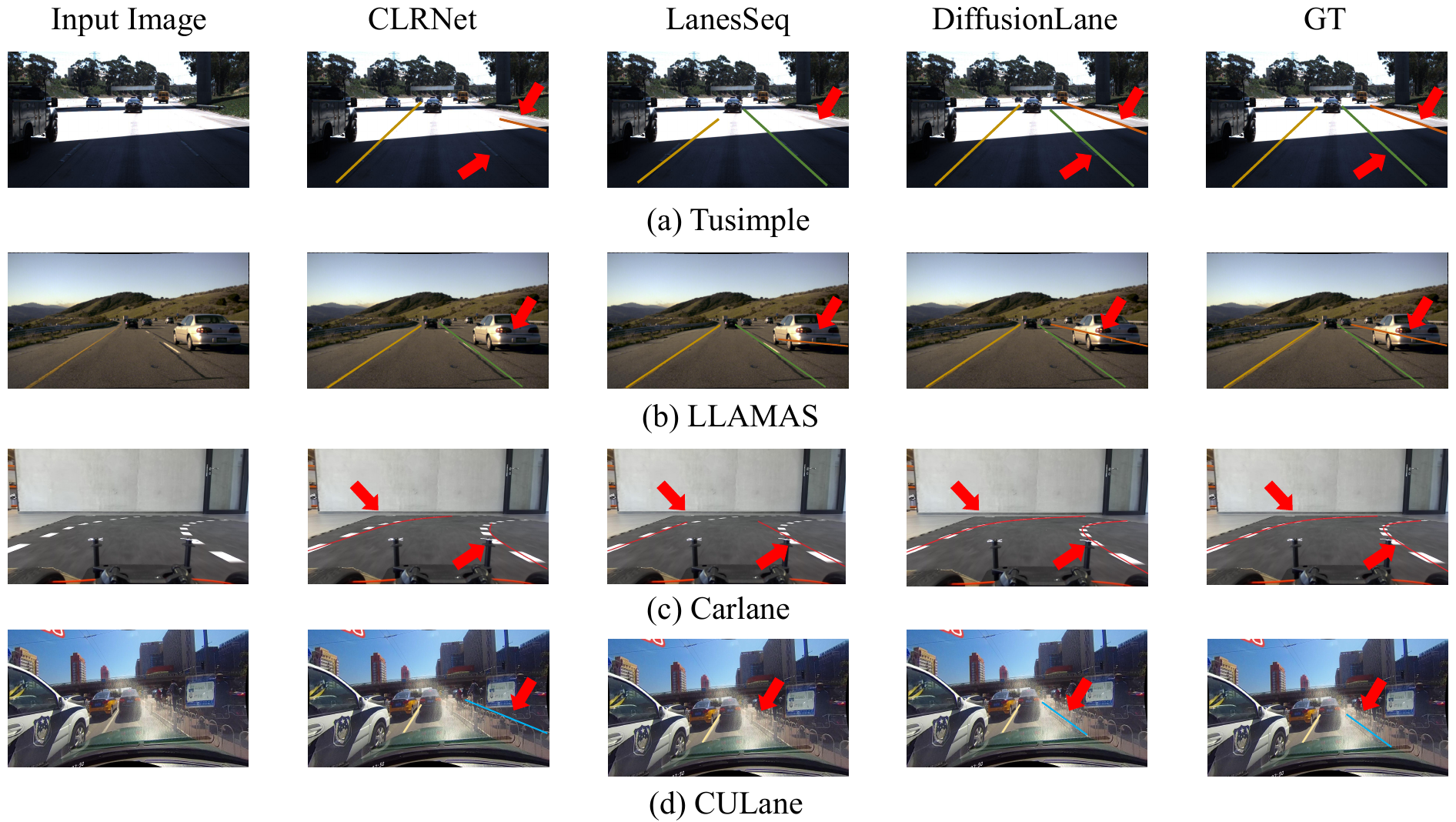}
        \caption{Visualization results of CLRNet, Lane2Seq, and DiffusionLane on four benchmarks.}
    \label{fig:8}
    % \vspace{-0.3cm}
\end{figure}

\textbf{Qualitative results}. We display the qualitative results in Figure~\ref{fig:8}. The results show that DiffusionLane can effectively detect lanes in the distribution-shift scenario (Figure~\ref{fig:8} (c)) but CLRNet and Lane2Seq does not.
Even in the extreme lightning scenario (Figure~\ref{fig:8} (a) and (d)), DiffusionLane is able to detect lanes successfully.

% \textbf{Qualitative results}. We display the qualitative results in Figure.

% \begin{figure}[t!]
%     \centering
%     \includegraphics[width=0.4\textwidth]{figure7.png}
%         \caption{The ablation study on the time step.}
%     \label{fig:7}
%     \vspace{-0.5cm}
% \end{figure}

% \begin{figure}[t!]
%     \centering
%     \includegraphics[width=0.4\textwidth]{figure7.png}
%         \caption{The ablation study on the time step.}
%     \label{fig:7}
%     % \vspace{-0.5cm}
% \end{figure}

\subsection{Ablation Study}
We conduct the ablation studies on the CULane dataset to evaluate the effectiveness of the key components in our method. We take CLRNet with ResNet34, SAFPN, and angle loss as the baseline. Additional ablation studies are provided in the supplementary materials.

% \begin{table}[t!]
%     \centering
%     \Large
%     \caption{Effectiveness of each component in DiffusionLane. CLRNet are adopted as the baseline.}
%     \resizebox{\columnwidth}{!}{
%         \begin{tabular}{cccccc}
%             \hline
%             Random lane anchors & Diffusion paradigm & Hybrid diffusion decoder & Auxiliary head & SAFPN& F1(\%) \\
%             \hline
%             & & & & & 79.73 \\
%             \hline
%             \checkmark & & & & & 74.39 \\
%             \checkmark & \checkmark & & & & 78.03 \\
%             \checkmark & \checkmark & \checkmark& & & 79.11 \\
%             \checkmark & \checkmark & \checkmark& \checkmark & & 79.89 \\
%             \checkmark & \checkmark & \checkmark& \checkmark & \checkmark& 80.24 \\
%             % & & & & 55.23 & 79.58 & 62.21 \
%             % \checkmark & & & & 55.61 & 79.81 & 62.76 \
%             % \checkmark & \checkmark & & & 55.61 & 80.03 & 63.09 \
%             % \checkmark & \checkmark & \checkmark & & 55.87 & 80.31 & \textbf{63.55} \
%             % \checkmark & \checkmark & \checkmark & \checkmark & 55.93 & \textbf{80.42} & 63.51 \
%             \hline
%             \end{tabular}
%     }
%     \label{tab:6}
% \end{table}

\textbf{Effectiveness of diffusion paradigm}. We first ablate the influence of the diffusion paradigm. As shown in Table~\ref{tab:6}, 
performance degrades from 79.96\% to 74.74\% when using random lane anchors, indicating that the quality of the lane anchors affects the performance significantly.
Equipped with our diffusion paradigm, F1 score is improved by 3.64\% (78.38\% vs. 74.74\%). The result manifests that the proposed diffusion paradigm has a 
good denoising effect. 

\textbf{Effectiveness of hybrid diffusion decoder}. Table~\ref{tab:6} shows that there still exists the quality gap between the lane anchors improved by the diffusion paradigm and the learnable lane anchors. Therefore,
we propose the hybrid diffusion decoder to further improve the quality of the lane anchors. We can see that hybrid diffusion decoder gains 1.08\% (79.46\% vs. 78.38\%) improvement in Table~\ref{tab:6}.
The effectiveness of hybrid diffusion decoder shows that 1) the feature refinement of lane anchors can improve the quality of lane anchors; 2) multi-decoders surpasses the single decoder by complementing each other.

\textbf{Effectiveness of the auxiliary head}. We further introduce an auxiliary head to improve the representation ability of the encoder. As presented in Table~\ref{tab:6}, the auxiliary head increases the F1 score 
from 79.46\% to 80.24\%. The result suggests that auxiliary positive samples generated by the auxiliary head is conductive to improving the performance via providing extra positive supervision singals.

\begin{table}[t!]
    \centering
    \Large
    \caption{Effectiveness of each component in DiffusionLane. RLA dentoes the random lane anchors.}
    \resizebox{\columnwidth}{!}{
        \begin{tabular}{cccccc}
            \hline
            RLA & Diffusion paradigm & Hybrid diffusion decoder & Auxiliary head & F1(\%) \\
            \hline
            & & & & 79.96 \\
            \hline
            \checkmark & & & & 74.74 \\
            \checkmark & \checkmark & & & 78.38 \\
            \checkmark & \checkmark & \checkmark& & 79.46 \\
            \checkmark & \checkmark & \checkmark& \checkmark & 80.24 \\
            % \checkmark & \checkmark & \checkmark& \checkmark & 80.24 \\
            % & & & & 55.23 & 79.58 & 62.21 \
            % \checkmark & & & & 55.61 & 79.81 & 62.76 \
            % \checkmark & \checkmark & & & 55.61 & 80.03 & 63.09 \
            % \checkmark & \checkmark & \checkmark & & 55.87 & 80.31 & \textbf{63.55} \
            % \checkmark & \checkmark & \checkmark & \checkmark & 55.93 & \textbf{80.42} & 63.51 \
            \hline
            \end{tabular}
    }
    \label{tab:6}
    % \vspace{-0.3cm}
\end{table}

% \begin{figure}[t!]
%     \centering
%     \includegraphics[width=0.5\textwidth]{figure7.png}
%         \caption{The ablation study on the time step.}
%     \label{fig:7}
% \end{figure}

% \begin{table}[t!]
%     \centering
%     \Large
%     \caption{Effectiveness of each component in DiffusionLane. CLRNet are adopted as the baseline.}
%     \resizebox{\columnwidth}{!}{
%         \begin{tabular}{cccccc}
%             \hline
%             Random lane anchors & Diffusion paradigm & Hybrid diffusion decoder & Auxiliary head & F1(\%) \\
%             \hline
%             & & & & 79.96 \\
%             \hline
%             \checkmark & & & & 74.74 \\
%             \checkmark & \checkmark & & & 78.38 \\
%             \checkmark & \checkmark & \checkmark& & 79.46 \\
%             \checkmark & \checkmark & \checkmark& \checkmark & 80.24 \\
%             % \checkmark & \checkmark & \checkmark& \checkmark & 80.24 \\
%             % & & & & 55.23 & 79.58 & 62.21 \
%             % \checkmark & & & & 55.61 & 79.81 & 62.76 \
%             % \checkmark & \checkmark & & & 55.61 & 80.03 & 63.09 \
%             % \checkmark & \checkmark & \checkmark & & 55.87 & 80.31 & \textbf{63.55} \
%             % \checkmark & \checkmark & \checkmark & \checkmark & 55.93 & \textbf{80.42} & 63.51 \
%             \hline
%             \end{tabular}
%     }
%     \label{tab:6}
% \end{table}

\textbf{Influence of the time step}. The denoising step in the inference stage can be viewed as the iterative denoising. The total time step $T$ decides the number of the iterations. As shown in Table~\ref{tab:7}, DiffusionLane achieves the best performance when $T=2$.
We find that when $T>2$, the model performance degrades, indicating that foreground lanes are filtered when $T$ is large. Besides, the larger time step brings more computational burden. Therefore, we set $T$ to 2.

% \begin{table}[t!]
%     \centering
%     % \Large
%     \small
%     \caption{The ablation study on the time step.}
%     \resizebox{\columnwidth}{!}{
%         \begin{tabular}{ccccc}
%             \hline
%              & $T=1$ & $T=2$ & $T=4$ & $T=6$ \\
%             \hline
%             F1 (\%) & 79.85 & \bf 80.24 & 79.98 & 79.62 \\
%             \hline
%             \end{tabular}
%     }
%     \label{tab:9}
%     % \vspace{-0.5cm}
% \end{table}

\begin{table}[t!]
    \centering
    % \Large
    % \tiny
    \caption{The ablation study on the sampling strategy.}
    \resizebox{\columnwidth}{!}{
        \begin{tabular}{cccccc}
            \hline
            DDIM & Lane anchors resampling & $T=1$ & $T=2$ & $T=4$ & $T=6$\\
            \hline
             &  & 77.63 & 77.35 & 76.98 & 76.11 \\
            \checkmark &  & 78.89 & 79.79 & 79.15 & 79.08 \\
             & \checkmark & 78.39 & 79.46 & 79.02 & 78.80 \\
            %  \rowcolor{gray!25}
            \checkmark & \checkmark & 79.85 & \bf 80.24 & 79.98 & 79.62 \\
            % \hline
            % & & & & & 79.73 \\
            % \hline
            % \checkmark & & & & & 74.39 \\
            % \checkmark & \checkmark & & & & 78.03 \\
            % \checkmark & \checkmark & \checkmark& & & 79.11 \\
            % \checkmark & \checkmark & \checkmark& \checkmark & & 79.89 \\
            % \checkmark & \checkmark & \checkmark& \checkmark & \checkmark& 80.24 \\
            \hline
            \end{tabular}
    }
    \label{tab:7}
\end{table}

\begin{table}[t!]
    \centering
    % \Large
    % \tiny
    \caption{The ablation study on the GT padding strategy.}
    \resizebox{0.95\columnwidth}{!}{
        \begin{tabular}{cccc}
            \hline
             & Repeating & Padding Gaussian & Padding Uniform \\
            \hline
            F1 (\%) & 78.27 & \bf 80.24 & 79.15 \\
            \hline
            \end{tabular}
    }
    \label{tab:8}
    % \vspace{-0.5cm}
\end{table}

\textbf{Sampling strategy}. We compare different sampling strategies in Table~\ref{tab:7}. We first evaluate DiffusionLane without DDIM, i.e., taking the output of the current step without the reverse chain as input for the next sampling step. Results show that DDIM can effectively improves the model performance. 
When equipped with lane anchors resampling, model performance gains the further enhancement, showing that DDIM and lane anchors resampling are beneficial.

\textbf{Ground truth padding strategy}. We study different padding strategy in Table~\ref{tab:8}. We consider three strategies: repeating the ground truth, padding the random lane anchors from the Gaussian distribution, and padding the random lane anchors from the uniform distribution. Results show that padding the random lane anchors from
the Gaussian distribution works best.

\section{Conclusion}
In this paper, we propose a novel lane detection paradigm, named DiffusionLane, casting the lane detection task as a diffusion process from the noisy lane anchors to target lanes.
Benefiting from the random lane anchors setting and denoising diffusion paradigm, DiffusionLane shows a strong generalization ability, enabling us to adopt the DiffusionLane in distribution-shift scenarios without re-training the model.
Moreover, we propose a hybrid decoding strategy, including the hybrid diffusion decoder and the auxiliary detection head, to achieve the better lane anchor refinement and feature representation. Experiments on four benchmarks 
demonstrate that DiffusionLane achieves favorable performance compared to the existing lane detectors.

\bibliography{aaai2026}

\clearpage

\begin{strip}
  \section{Suppmentary Material for DiffusionLane: Diffusion Model for Lane Detection}
\end{strip}

% \onecolumn
% % --- 附录大标题（单栏居中） ---
% \centering
% {\LARGE \bfseries Suppmentary Material for DiffusionLane: Diffusion Model for Lane Detection}\\[0.5em]
% % -------------------------------
% % ===== 恢复双栏，进入附录 =====
% \twocolumn[]
\appendix

\begin{abstract}
In this suppmentary material, we first provide the pseudo-codes of the training and the inference stage. Then, we supplemet the additional experiments, including the 
evaluation metrics, additional implementation details, and additional ablation studies. Finally, we describe the limitation and broader impact.
\end{abstract}

\section{Method}
\subsection{Pseudo-codes of training and inference stage}

The pseudo-codes of the training and inference stage are provided in Algorithm~\ref{alg:train} and ~\ref{alg:sample}. In Algorithm~\ref{alg:train}, \texttt{alpha\_cumprod(t)} = $\prod_{i=1}^t \bar{\alpha}_i$. 
In Algorithm~\ref{alg:sample}, \texttt{linespace} means generating evenly spaced values.

\begin{algorithm}[h]
\caption{Training 
}
\label{alg:train}
% \algcomment{\fontsize{7.2pt}{0em}\selectfont \texttt{alpha\_cumprod(t)}: cumulative product of $\alpha_i$, i.e., $\prod_{i=1}^t \alpha_i$
% }
\definecolor{codeblue}{rgb}{0.25,0.5,0.5}
\definecolor{codegreen}{rgb}{0,0.6,0}
\definecolor{codekw}{RGB}{207,33,46}
\lstset{
    backgroundcolor=\color{white},
    basicstyle=\fontsize{7.5pt}{7.5pt}\ttfamily\selectfont,
    columns=fullflexible,
    breaklines=true,
    captionpos=b,
    commentstyle=\fontsize{7.5pt}{7.5pt}\color{codegreen},
    keywordstyle=\fontsize{7.5pt}{7.5pt}\color{codekw},
    escapechar={|}, 
}
\begin{lstlisting}[language=python]
    def train_loss(images, gt_lanes):
      """
      images: [B, H, W, 3]
      gt_lanes: [B, *, 76]
      # B: batch
      # N_train: number of lane anchors
      # 76=72+4, where 72 is the number of offsets and 4 represent the x,y coordinates of the starting point, the angle, and the length
      """
      
      # Encode image features
      feats = encoder(images)
      
      # Pad gt_lanes to N_train
      pb = pad_boxes(gt_lanes) # padded boxes: [B, N_train, 76]

      # Construct the target lane anchors
      pb = gt_to_anchor(pb) # [B, N_train, 76]

      # select the starting point and the angle as the diffusion target
      diff_target = pd[:,:,:3]
    
      # Normalize
      diff_target = (diff_target * 2 - 1) * scale  
    
      # Corrupt gt_lanes
      t = randint(0, T)|~~~~~~~~~~~|# time step
      eps = normal(mean=0, std=1)  # noise: [B, N_train, 3]
      pb_crpt = sqrt(|~~~~|alpha_cumprod(t)) * diff_target + 
                  |~|sqrt(1 - alpha_cumprod(t)) * eps
    
      # construct noisy lane anchors
      pb_crpt = point_to_anchor(pb_crpt) # [B, N_train, 76]

      # Predict
      pb_pred = hybrid_diffusion_decoder(pb_crpt, feats, t)
    
      # Compute loss
      loss = compute_loss(pb_pred, gt_lanes)
      loss_aux = aux_head(pb_pred, gt_lanes)
      loss = loss + loss_aux
      
      return loss
    \end{lstlisting}
\end{algorithm}

\begin{algorithm}[h]
    \caption{DiffusionLane Sampling 
    }
    \label{alg:sample}
    % \algcomment{\fontsize{7.2pt}{0em}\selectfont \texttt{linespace}: generate evenly spaced values
    % }
    \definecolor{codeblue}{rgb}{0.25,0.5,0.5}
    \definecolor{codegreen}{rgb}{0,0.6,0}
    \definecolor{codekw}{rgb}{0.85, 0.18, 0.50}
    \lstset{
      backgroundcolor=\color{white},
      basicstyle=\fontsize{7.5pt}{7.5pt}\ttfamily\selectfont,
      columns=fullflexible,
      breaklines=true,
      captionpos=b,
      commentstyle=\fontsize{7.5pt}{7.5pt}\color{codegreen},
      keywordstyle=\fontsize{7.5pt}{7.5pt}\color{codekw},
      escapechar={|}, 
    }
    \begin{lstlisting}[language=python]
    def infer(images, steps, T):
      """
      images: [B, H, W, 3]
      # steps: number of sample steps
      # T: number of time steps
      """
      
      # Encode image features
      feats = encoder(images)
      
      # sample the starting point and angle
      # pb_t: [B, N_train, 3]
      pb_t = normal(mean=0, std=1)

      # construct noisy lane anchors
      pb_t = point_to_anchor(pb_t)
    
      # uniform sample step size
      times = reversed(linespace(-1, T, steps))
      
      # [(T-1, T-2), (T-2, T-3), ..., (1, 0), (0, -1)]
      time_pairs = list(zip(times[:-1], times[1:])
    
      for t_now, t_next in zip(time_pairs):
        # Predict pb_0 from pb_t
        pb_pred = hybrid_diffusion_decoder(pb_t, feats, t_now)
        
        # Estimate pb_t at t_next
        pb_t = ddim_step(pb_t, pb_pred, t_now, t_next)
    
        # Lane anchor resampling
        pb_t = lane_anchor_resampling(pb_t)
        
      return pb_pred
    \end{lstlisting}
\end{algorithm}

\section{Additional Experiments}

\subsection{Evaluation metrics}

For CULane and LLAMAS dataset, we utilize F1 score to measure the performance: $F_1 = \frac{2 \times Precision \times Recall}{Precision + Recall}$, where $Precision=\frac{TP}{TP+FP}$ and $Recall=\frac{TP}{TP+FN}$. $TP$, $FP$, and $FN$ represent the true positive rate, the false positive rate, and the false negative rate, respectively.

For Tusimple and Carlane datasets, we adopt accuracy, FP, and FN to evaluate the model performance. Accuracy is defined as $Accuracy = \frac{\sum_{clip} C_{clip}}{\sum_{clip} S_{clip}}$, $C_{clip}$ denotes the number of accurately predicted lane points and $S_{clip}$ represents the total number of lane points of a clip. A lane point is treated as a correct point if its distance is smaller than the threshold $t_{pc} = \frac{20}{cos(a_{yl})}$, here $a_{ul}$
is the angle of the corresponding ground truth.

\subsection{Additional implementation details}
We select the ResNet~\cite{he2016deep}, DLA~\cite{yu2018deep}, and MobileNetV4-Hybrid-M~\cite{qin2024mobilenetv4} as the image encoder and all image encoders are initialized with pretrained weights on ImageNet1K~\cite{deng2009imagenet}. We train DiffusionLane using AdamW optimizer with learning rate 0.0003. Cosine schedule is adopted to adjust the learning rate. All models are trained on a single 3090 GPU with 24 GB memory and batch size is 20. 
The training epoches are set to 70,25,20,20 for Tusimple, CULane, LLAMAS, and Carlane datasets, respectively. The number of lanes $N_{train}$ in the ground truth padding is 800.
We set $Y_{min}$ to 160, 270, 300, and 160 for Tusimple, CULane, LLAMAS, and Carlane.

\subsection{Additional ablation studies}
In this section, we provide the additional ablation studies. If not specified, all supplemeted experiments are conducted on CULane and image encoder is ResNet34.

\begin{table}[h!]
    \centering
    % \Large
    \tiny
    \caption{The ablation study on the noise scale.}
    \resizebox{\columnwidth}{!}{
        \begin{tabular}{cccccc}
            \hline
             & 0.1 & 0.5 & 1 & 2 & 3 \\
            \hline
            F1 (\%) & 77.35 & 78.69 & 79.42 & \bf 80.24 & 80.11 \\
            \hline
            \end{tabular}
    }
    \label{tab:1}
\end{table}

\textbf{Noise scale}. We first ablate the influence of the nosie scale and the results are presented on Table~\ref{tab:1}. We can see that the model achieves the optimal performance when 
the noise scale is 2, which is higher than the image generation task (noise scale is 1)~\cite{song2020denoising} and segmentation task (noise scale is 0.1)~\cite{ji2023ddp}.

\begin{table}[h!]
    \centering
    % \Large
    \tiny
    \caption{The ablation study on the number of lane anchors $N_{train}$.}
    \resizebox{\columnwidth}{!}{
        \begin{tabular}{cccccc}
            \hline
             & 192 & 384 & 600 & 800 & 1000 \\
            \hline
            F1 (\%) & 76.11 & 77.53 & 78.90 & \bf 80.24 & 80.31 \\
            \hline
            \end{tabular}
    }
    \label{tab:2}
\end{table}

\textbf{The number of lane anchors}. We study the number of lane anchors $N_{train}$. Results in Table~\ref{tab:2} show that DiffusionLane prefers a high $N_{train}$ (800 lane anchors).
We explain that if $N_{train}$ is small, the lane anchors around the target lanes are sparse due to the uniform distribution. This leads to hard optimization during the training compared
to the learnable lane anchors. However, when $N_{train}$ is large enough, the computational burden increases. Considering the performance and the efficiency, we set $N_{train}$ to 800.

\begin{table}[h!]
    \centering
    % \Large
    \tiny
    \caption{The ablation study on FPS.}
    \resizebox{\columnwidth}{!}{
        \begin{tabular}{cccccc}
            \hline
             & 192 & 384 & 600 & 800 & 1000 \\
            \hline
            FPS & 88 & 72 & 65 & 52 & 39 \\
            \hline
            \end{tabular}
    }
    \label{tab:3}
\end{table}

\textbf{Analysis on the FPS}. Table~\ref{tab:3} presents the study on FPS. We test te inference speed on a single 2080Ti GPU without TensorRT. It can be observed that DiffusionLane does not have the advantages in the inference speed.
Reasons may be that DiffusionLane requires the multiple runs on the decoder, which affects the inference speed. We think this is a limitation of DiffusionLane and our ongoing work is improving the 
inference speed.

\begin{table}[h!]
    \centering
    % \small
    \caption{Transferring the hybrid decoding technology to the existing lane detectors.}
    \resizebox{\columnwidth}{!}{
        \begin{tabular}{lc c}
            \hline
            Methods& Image encoder &F1(\%)$\uparrow$ \\
            \hline
            CLRNet~\cite{zheng2022clrnet} & ResNet34 & 79.73 \\
            CLRNet+hybrid decoding & ResNet34 & 80.58 \\
            CLRerNet~\cite{honda2024clrernet} & ResNet34 & 80.67 \\
            CLRerNet+hybrid decoding & ResNet34 & 81.11 \\
            ADNet~\cite{xiao2023adnet} & ResNet34 & 78.94 \\
            ADNet+hybrid decoding & ResNet34 & 79.79 \\
            % PolyLaneNet~\cite{tabelini2021polylanenet} & EfficientNet-B0~\cite{tan2019efficientnet} & 90.20 \\
            % BezierLaneNet~\cite{feng2022rethinking} & ResNet34 & 96.11 \\
            % LaneATT~\cite{tabelini2021keep} & ResNet34 & 94.96 \\
            % LaneATT~\cite{tabelini2021keep} & ResNet122 & 95.17 \\
            % LaneAF~\cite{abualsaud2021laneaf} & DLA34 & 96.90 \\
            % CLRNet~\cite{zheng2022clrnet} & ResNet18 & 96.96 \\
            % CLRNet~\cite{zheng2022clrnet} & DLA34 & 97.16 \\
            % Lane2Seq (segmentation)~\cite{zhou2024lane2seq} & ViT-Base~\cite{dosovitskiy2020image} & 97.42 \\
            % \hline
            % \rowcolor{gray!25}
            % DiffusionLane & ResNet18 & 97.27 \\
            % \rowcolor{gray!25}
            % DiffusionLane & ResNet34 & 97.36 \\
            % \rowcolor{gray!25}
            % DiffusionLane & ResNet101 & \bf 97.59 \\
            \hline
        \end{tabular}
    }
    \label{tab:4}
\end{table}

\textbf{Transferring hybrid decoding to other lane detectors}. Table~\ref{tab:4} presents the results on transferring the hybrid decoding technology to the existing anchor-based lane detectors.
Results demonstrate that our hybrid decoding can brings the consistent improvement on the existing anchor-based methods, indicating that hybrid decoding technology can help optimize the 
parameters of the learnable lane anchors.

\begin{table}[h!]
    \centering
    % \small
    \caption{Comparison of different diffusion targets.}
    \resizebox{\columnwidth}{!}{
        \begin{tabular}{lc c}
            \hline
            Diffusion target  &F1(\%)$\uparrow$ & Training epoches\\
            \hline
            Starting point and angle & 80.24 & 25 \\
            All lane points & 78.95 & 25 \\
            All lane points & 80.21 & 40 \\
            % PolyLaneNet~\cite{tabelini2021polylanenet} & EfficientNet-B0~\cite{tan2019efficientnet} & 90.20 \\
            % BezierLaneNet~\cite{feng2022rethinking} & ResNet34 & 96.11 \\
            % LaneATT~\cite{tabelini2021keep} & ResNet34 & 94.96 \\
            % LaneATT~\cite{tabelini2021keep} & ResNet122 & 95.17 \\
            % LaneAF~\cite{abualsaud2021laneaf} & DLA34 & 96.90 \\
            % CLRNet~\cite{zheng2022clrnet} & ResNet18 & 96.96 \\
            % CLRNet~\cite{zheng2022clrnet} & DLA34 & 97.16 \\
            % Lane2Seq (segmentation)~\cite{zhou2024lane2seq} & ViT-Base~\cite{dosovitskiy2020image} & 97.42 \\
            % \hline
            % \rowcolor{gray!25}
            % DiffusionLane & ResNet18 & 97.27 \\
            % \rowcolor{gray!25}
            % DiffusionLane & ResNet34 & 97.36 \\
            % \rowcolor{gray!25}
            % DiffusionLane & ResNet101 & \bf 97.59 \\
            \hline
        \end{tabular}
    }
    \label{tab:5}
\end{table}

\textbf{Diffusion target}. We select the starting point coordinate and the angle of the lane anchor as the diffusion target. An alternative diffusion paradigm is adding the Gaussian noise to all points
of a lane anchor. We compare the above two methods in Table~\ref{tab:5}. The results show that adopting all lane points as the diffusion target require more training epoches than the starting point and the angle,
indicating that selecting all lane points as the diffusion target leads to hard optimization. Besides, diffusing all lane points brings extra computational burden. Therefore, we adopt the starting point and the angle as the diffusion target.

\section{Limitation and Broader Impact}
\subsection{Limitation}
As discussed in the Table~\ref{tab:3}, one major limitation of DiffusionLane is the inference speed. Reasons lie in two folds: 1) dense lane anchors. 2) multiple runs of the decoder. In the further, we will focus on developing a decoding technology with high efficiency.

\subsection{Broader Impact}
From the results presented by DiffusionLane, we believe the diffusion model is a new possible model for lane detection due to its random lane anchors setting and denoising diffusion paradigm, showing a strong generalization ability on the distribution-shift scenarios without retraining the model.
We hope our method can serve as a simple yet effective baseline for anchor-based methods and inspire designing more high-performance and efficient architecture.

\end{document}